\documentclass{article}

 \PassOptionsToPackage{numbers}{natbib}
 \usepackage[preprint]{neurips_2026}

\usepackage[utf8]{inputenc} % allow utf-8 input
\usepackage[T1]{fontenc}    % use 8-bit T1 fonts
\usepackage{hyperref}       % hyperlinks
\usepackage{url}            % simple URL typesetting
\usepackage{booktabs}       % professional-quality tables
\usepackage{amsfonts}       % blackboard math symbols
\usepackage{nicefrac}       % compact symbols for 1/2, etc.
\usepackage{microtype}      % microtypography
\usepackage{xcolor}         % colors

\usepackage{graphicx}
\usepackage{multirow}
\usepackage{colortbl}
\usepackage{wrapfig}
\usepackage{enumitem}
\usepackage{array}
\usepackage{float}
\usepackage{subcaption}
\usepackage{xcolor}
\usepackage{bbding}
\usepackage{makecell}
\usepackage{caption}
\usepackage{amsmath}
\usepackage{wasysym}

\title{The Illusion of Forgetting: Attack Unlearned Diffusion \\ via Initial Latent Variable Optimization}

\author{%
  Manyi Li \\
  School of Advanced Interdisciplinary Sciences\\
  University of the Chinese Academy of Sciences\\
  Beijing, China \\
    \And
  Yufan Liu \\
  State Key Laboratory of Multimodal Artificial Intelligence Systems \\
  Institute of Automation \\
  Chinese Academy of Sciences \\
  Beijing, China \\
   \AND
    Lai Jiang \\
  Electronic Engineering \\
  Beijing University of Aeronautics and Astronautics \\
  Beijing, China \\
  \AND
  Bing Li \thanks{Corresponding Author} \\
  State Key Laboratory of Multimodal Artificial Intelligence Systems \\
  Institute of Automation \\
  Chinese Academy of Sciences \\
  Beijing, China \\
  \AND
  Yuming Li \\
  Alipay \\
  Ant Group \\
  \AND
  Weiming Hu \\
  State Key Laboratory of Multimodal Artificial Intelligence Systems \\
  Institute of Automation \\
  Chinese Academy of Sciences \\
  Beijing, China \\
}

\author{%
Manyi Li$^{1}$\quad Yufan Liu$^{2,3}$ \quad Lai Jiang$^{1}$\quad Bing Li$^{2,3}$ \thanks{Corresponding Author} \quad  Yuming Li$^{4}$ \quad Weiming Hu$^{2,3}$\\
$^{1}$ School of Advanced Interdisciplinary Sciences, University of the Chinese Academy of Sciences \\
$^2$ State Key Laboratory of Multimodal Artificial Intelligence Systems \\
$^3$ Institute of Automation, Chinese Academy of Sciences\\
$^4$ Beijing University of Aeronautics and Astronautics \\
$^5$ Alipay, Ant Group \\
}

\begin{document}

\maketitle

\begin{abstract}
Text-to-image diffusion models (DMs) are frequently abused to produce harmful or copyrighted content, violating public interests. Concept erasure (unlearning) is a promising paradigm to alleviate this issue. However, there exists a peculiar forgetting illusion phenomenon with unclear cause. Based on empirical analysis, we formally explain this cause: most unlearning partially disrupt the mapping between linguistic symbols and the underlying internal knowledge, leaving the knowledge intact as \textbf{dormant memories}. We further demonstrate that distributional discrepancy in the denoising process serves as a measurable indicator of how much of the mapping is retained, also reflecting unlearning strength. Inspired by this, we propose \textbf{IVO} (\textbf{I}nitial Latent \textbf{V}ariable \textbf{O}ptimization), a novel attack framework designed to assess the robustness of current unlearning methods. IVO optimizes initial latent variables to realign the noise distribution of unlearned models with that of their vanilla counterparts, which reconstructs the fractured mappings and consequently revives dormant memories. Extensive experiments covering 11 unlearning techniques and 3 concept scenarios show that IVO outperforms state-of-the-art baselines, exposing fundamental flaws in current unlearning mechanisms. 
%The code is available at \url{anonymous.4open.science/r/IVO/}. 
\textbf{Warning}: This paper has unsafe images that may offend some readers.
\end{abstract}

\section{Introduction} \label{introduction}
The rapid advancement of Diffusion Models (DMs) has revolutionized image generation, enabling the creation of realistic imagery from simple textual descriptions. However, this power is a double-edged sword, as DMs can be exploited to produce Not Safe For Work (NSFW) and copyrighted content. Concept erasure, termed ``unlearning'', is a promising technique to mitigate this problem. Unlearning removes undesirable concepts from the model itself while preserving its general performance.

\begin{figure*}[!t]
\centering
\includegraphics[width=\textwidth]{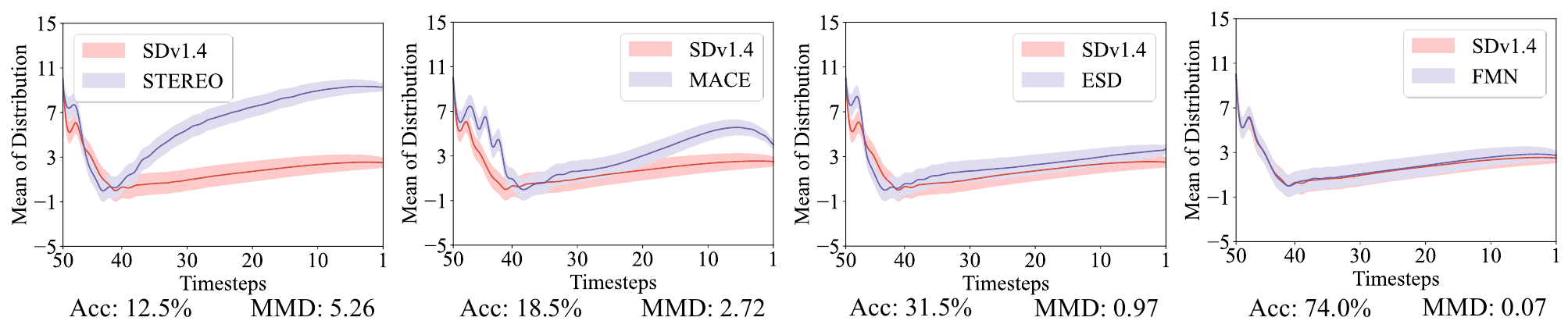}
\caption{ The non-negligible Acc indicates that unlearned DMs partially destroy symbol-to-knowledge mappings of the target concept. The MMD value measures a discrepancy between two distribution trajectories of predicted noise, reflecting the damage extent of the mappings. SDv1.4 is a vanilla DM used for reference.}
\label{Unlearn_capability}
\vspace{-10pt}
\end{figure*}

% 直接将提示词输入unlearned diffusion，仍有一定比例的生成图片含有目标概念，说明这一种局部最优。
% method 第一段解释ACC怎么来的
According to the non-negligible detection accuracy (Acc) in Fig.~\ref{Unlearn_capability}, most unlearning methods actually achieve a local optimum. This is an illusion of ``forgetting'' also confirmed by prior work \cite{suriyakumar2024unstable, rusanovsky2025memories}, but a plausible interpretation of its cause remains absent. We model the generation of images containing specific concepts as a mapping process between linguistic symbols (e.g., Van Gogh) and underlying knowledge (e.g., the corresponding visual representational features, structural characteristics and attribute correlations such as contours, textures, spatial layouts). Destroying this process requires either eliminating the knowledge or disrupting \textit{the symbol-to-knowledge mapping pathways}. Consequently, three important conclusions can further be derived: \textbf{(1) Unlearning does not destroy knowledge stored in model's parameters.} If the knowledge were truly eradicated, the mapping would fail to reproduce the complete concept, resulting in detection failure. However, the non-negligible Acc contradicts this, confirming that the knowledge remains intact as \textbf{dormant memories}. \textbf{(2) the illusion of ``forgetting'' comes from incomplete mapping deconstruction.} The first conclusion indicates that unlearning disrupts the mapping pathways, but the non-negligible Acc demonstrates that there still exist valid mappings despite much less than before. \textbf{(3) This reduction in mapping pathways positively correlates with distributional discrepancy.} As shown in Fig.~\ref{Unlearn_capability}, a larger divergence between unlearned and vanilla DMs is associated with stronger unlearning (fewer mapping pathways). The first two conclusions provide a theoretical interpretation for the illusion of ``forgetting'' phenomenon, while the last implies that dormant memories can be reactivated by reducing the distributional gap.

Given these insights, we propose \textbf{IVO} (\textbf{I}nitial Latent \textbf{V}ariant \textbf{O}ptimization), an attack framework that assesses adversarial robustness of unlearned DMs.
Unlike prior work that relies on prompt engineering \cite{tsai2024ring, chin2024prompting4debugging}, IVO uses initial latent variables as triggers, operating directly in the image latent space where unlearning paradigm has less influence. This enables more effective and semantically consistent concept reactivation. Specifically, IVO firstly uses DDIM inversion to map a reference image into latent space and takes it as the initial latent variable. This provides a both directionally and semantically aligned starting point that enables faster convergence in the broad optimization space. This latent is refined via a dual-loss objective. A distribution matching loss (DML) aligns the noise distribution of the unlearned DM with that of a vanilla DM, effectively reconstructing the broken symbol-to-knowledge mapping. A direction calibration loss (DCL) strictly anchor the optimization toward the semantic manifold of the erased concept to avoid semantic drift. IVO stores successful latents in a pool and reuses them during subsequent attacks, reducing the heavy computational burden. To reveal flaws in existing unlearning methods and verify the effectiveness of IVO, we conduct comprehensive evaluation on eleven widely used unlearning methods across three representative concept scenarios. IVO achieves superior attack performance and semantic fidelity in terms of baselines. Our contributions are summarized as follows: 
\begin{itemize}[leftmargin=10pt]
\item We theoretically interpret the fundamental cause of the illusion of ``forgetting'' phenomenon. It is because most unlearning methods only partially disrupt \textit{the symbol-to-knowledge mapping}, while leaving knowledge intact as \textbf{dormant memories}. 
\item We first demonstrate that distributional discrepancy positively correlates with unlearning strength. Inspired by this finding, we propose IVO, a novel attack framework that allows unlearned DMs to generate erased concepts by optimizing initial latent variables.
\item Extensive experiments thoroughly validate the effectiveness of IVO involving 11 popular unlearning methods across 3 representative scenarios. IVO shows superior attack performance and semantic consistency compared to attack baselines.
\end{itemize}

\section{Related work}

\subsection{Concept Erasure}
Concept erasure, termed ``unlearning,'' is designed to eliminate certain undesirable concepts that a model has learned, including copyrighted content and pornographic material. ESD \cite{gandikota2023erasing}, SLD \cite{schramowski2023safe} and UCE \cite{gandikota2024unified} are pioneering works, representing three mainstreams. ESD fine-tunes a pretrained model using only the target concept name, achieving specific visual concept unlearning. In contrast, SLD directly use a negative guidance algorithm to manipulates predicted noise in inference. UCE employs a closed-form solution to edit parameters without fine-tuning. 
However, ``unlearning'' inevitably affects normal generation. Consequently, several efforts \cite{SFD, AGE} have focused on balancing concept removal with preserving normal generation. AdvU \cite{zhang2025defensive} and MetaU \cite{Meta} introduce different objectives to improve robustness against adversarial attacks and fine-tuning, respectively. 

%Other studies \cite{ren2024unveiling, rusanovsky2025memories} reveal that concept memory persist in specialized model components rather than being fully erased.
\subsection{Safety Evaluation and Analysis of Unlearned DM}

Several studies have developed dedicated jailbreaking pipelines to quantify the safety of unlearned DMs. P4D \cite {chin2024prompting4debugging} is a white-box red-teaming tool that crafts gradient-guided adversarial texts to circumvent unlearning defenses. Conversely, Ring \cite {tsai2024ring} is a model-agnostic black-box evaluation framework that produces transferable prompts via latent concept representation extraction. UDiff \cite {zhang2024generate} taps into DMs’ intrinsic classification capability to build attack prompts with no dependency on auxiliary models. Recall \cite{liu2025image} conducts a multi-modal adversarial attack via optimizing image prompts with fixed text inputs to reactivate erased concepts while preserving semantic alignment. SubAttack \cite{chen2025dual} learns interpretable orthogonal token embeddings to expose implicit semantic associations of erased concepts in unlearned DMs.

By special experiments, some work analyzed potential flaws of unlearning methods. \citet{suriyakumar2024unstable} confirmed closed-form unlearning is far more resurgence-robust than fine-tuning-based approaches. \citet{luconcepts}
defines trajectories as latent sampling and concept evolution paths, and analyze them by similarity of CLIP embedding, differentiating two concept erasure paradigms. It is distinct with ours noise distribution trajectories.
Different with we explicitly formulates dormant memory as the underlying knowledge of concept definition of dormant memory, \citet{rusanovsky2025memories} vaguely defines it as ``concept information''. \citet{rusanovsky2025memories} found erased concept information persists in DMs’ latent space and is reconstructible via diffusion inversion-retrieved latent seeds, but depends on a strict prerequisite: generating an image $\mathcal{I}$ for a given text requires prior access to $\mathcal{I}$ itself. \citet{george2025illusion} uncovered inherent instability of current unlearning methods that erased content revives with benign fine-tuning on unrelated concepts.

\section{Preliminary}
\subsection{Diffusion Model}
Diffusion Models (DMs) are dominant deep generative models for high-fidelity data synthesis, built on an iterative denoising paradigm. In the training phase, a fixed forward diffusion process progressively corrupts clean data $z_0$ into noised samples $z_t$, defined as $\displaystyle q(z_t \mid z_0) = \mathcal{N}\left(z_t; \sqrt{\bar{\alpha}_t} z_0, (1-\bar{\alpha}_t) \mathbf{I}\right)$, where $\bar{\alpha}_t$ is the cumulative product of the pre-defined noise schedule. A neural network $\epsilon_\theta$ is then trained to predict the injected noise by minimizing the objective $\displaystyle \mathcal{L}_{\text{DM}} = \mathbb{E}_{z_0, t, \epsilon \sim \mathcal{N}(0,\mathbf{I})} \left[ \left\| \epsilon - \epsilon_\theta(z_t, t) \right\|_2^2 \right]$. In the inference phase, the model iteratively refines a random-initialized Gaussian noise $z_T$ by $\displaystyle p_\theta(z_{t-1} \mid z_t) = \mathcal{N}\left(z_{t-1}; \mu_\theta(z_t, t), \sigma_t^2 \mathbf{I}\right)$, where $\mu_\theta(z_t, t)$ is the network-predicted posterior mean and $\sigma_t$ controls the generation stochasticity.

\subsection{Problem Formulation} \label{problem formulation}
We primarily focus on evaluating the adversarial robustness of unlearned DMs in the white-box setting, where the unlearned DM is full accessible to enable end-to-end adversarial optimization. Formally, let $\theta_c$ denote a DM that has erased the concept $c$ via unlearning techniques and failed to generate recognizable $c$-related content when given a prompt $P_c$ explicitly containing $c$. We formulate the core problem of this work as follows:
\begin{itemize}[leftmargin=*]
    \item \textbf{Concept Reactivation}. Jailbreak the unlearning defense of $\theta_c$, enabling the model to generate images with clearly identifiable $c$-related content via our crafted adversarial input $\hat{z}_{adv}^T$. We quantify this goal by the Attack Success Rate (ASR).
    \item \textbf{Semantic Consistency}. Preserve high semantic alignment between generated image and the original prompt $P_c$, eliminating semantic drift. This constraint ensures the high ASR of attack is not meaningless, and guarantees the attack can reveal exploitable security vulnerabilities of the unlearned DMs. We evaluate this by CLIP score (CLIP) and Kernel Inception Distance (KID).
\end{itemize}

\begin{figure*}[!t]
\centering
\includegraphics[width=\textwidth]{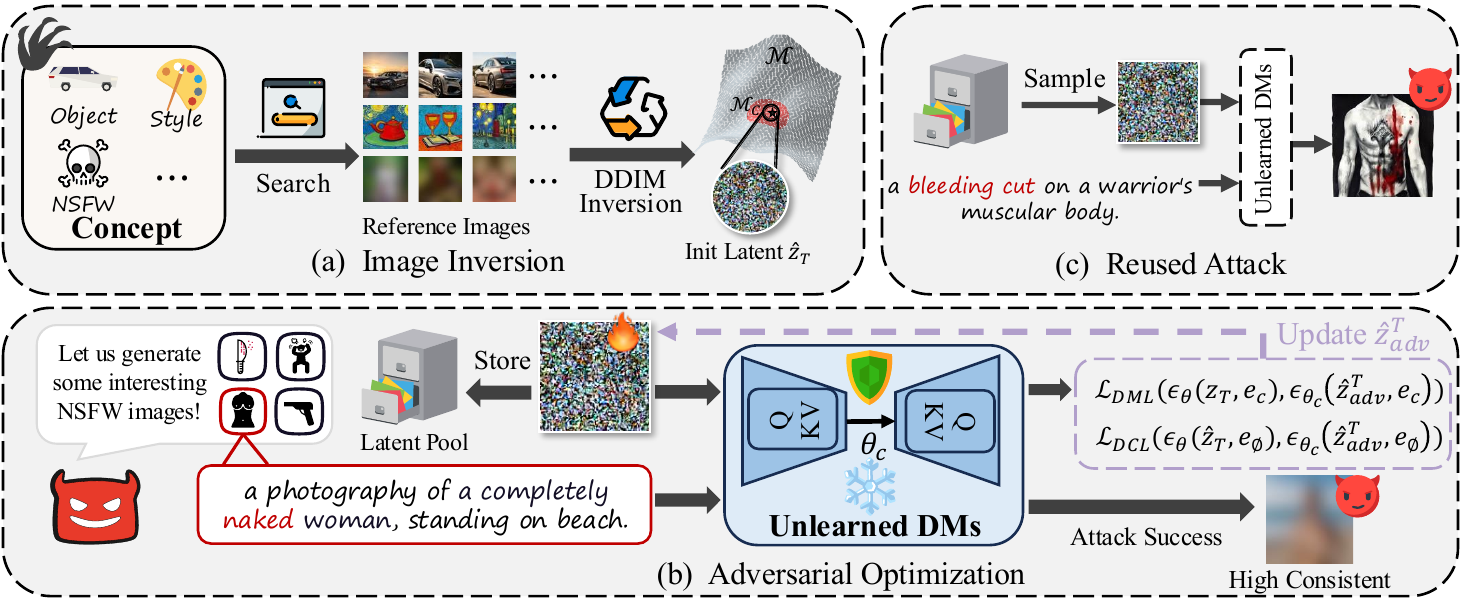}
\caption{Overview of the proposed IVO attack framework.}
\label{attack_framework}
\vspace{-10pt}
\end{figure*}

\section{Method} \label{sec:method}
\subsection{Framework Overview}
As illustrated in Fig.~\ref{attack_framework}, IVO optimizes the initial latents in the image latent space to revive erased concepts in unlearned DMs, rather than tuning text prompts. It first uses DDIM inversion to convert a reference image $\mathcal{I}_c$ containing $c$-related content into semantically aligned initial latent $\hat{z}_T$, and then optimizes it via a dual-loss objective for denoising distribution alignment and semantic preservation. Finally effective $\hat{z}_{adv}^T$ are stored in a pool for efficient reuse in subsequent attacks.

\subsection{Motivation of the IVO Framework}
To conduct experiments in Fig.~\ref{Unlearn_capability}, we feed numerous prompts $P_c$ into unlearned DMs for image generation, and then examine the conceptual content in the generated images via detectors. The results show that unlearned DMs still produce a considerable proportion of images $\mathcal{I}_c$. Based on analysis in Section \ref{introduction}, this reveals that these models are, in fact, unable to completely remove the target concept $c$, and retain part of the symbol-to-knowledge mappings. The extent of such retention is positively correlated with distributional discrepancy, which offers new insights for measuring unlearning effectiveness. Specifically, vanilla and unlearned DMs generate images independently using a dataset $\mathcal{D}$ with more than 500 $P_c$. We record the predicted noise distribution at each inference step, then average the mean and variance across the dataset respectively. We then visualize the distribution trajectories and compute the Maximum Mean Discrepancy (MMD) \cite{gretton2012kernel} between them. As shown in Fig. \ref{Unlearn_capability}, curves with higher alignment yield smaller MMD values, implying more valid mapping pathways and poorer unlearning ability. For instance, FMN achieves superior curve-fitting with the lowest MMD (0.07) and an Acc of 74.0\%. Conversely, STEREO presents a large discrepancy with 12.5\% Acc and the highest MMD (5.26). This correlation suggests that it is feasible to reconstruct the disrupted symbol-to-knowledge mapping by realigning noise distribution trajectories. 

\subsection{Rationale of Initial Latent Optimization}
Consider an erased concept $c$ and its related concept $c^{\ast}$, along with their corresponding symbol-to-knowledge mappings $\mathcal{F}$ and $\mathcal{F}^{\ast}$. When it comes to the removal of $c$, existing unlearning techniques inevitably exert an adversarial impact on $\mathcal{F}^{\ast}$. Unfortunately, regardless of whether prompt-based attacks stem from vocabulary-level (replacing sensitive words) or syntactic perturbation (injecting trainable prefixes), they share a common characteristic: \textit{searching for concepts similar to the erased concept within the text semantic space}. Consequently, the inherent commonality of prompt-based attacks constrains their effectiveness against unlearned DMs, because the mapping of related concepts $\mathcal{F}^{\ast}$ is also compromised in the unlearning process. To break the inherent limitations of prompt-based attacks, we select the initial latent $z_T$ as our optimization object. $z_T$ resides in the image latent space, which offers rich and extensive search pathways for reconstruction, and it is another necessary input condition for DMs, indicating broad applicable scenarios.

\subsection{Optimization Objective Formulation}
Based on preceding insights, we first formulate a straightforward constrained optimization goal to address problems in Section \ref{problem formulation}, and then derive its dual-loss form used by IVO via rigorous proof.

Our primal goal is twofold but encoded in a single constrained optimization problem: \textbf{(a)} to align the conditional generative distribution of the unlearned model $\theta_c$ with that of the surrogate model $\theta$ to reconstruct the broken symbol-to-knowledge mapping; \textbf{(b)} to strictly anchor the adversarial latent $z_{adv}^T$ to the semantic manifold $\mathcal{M}_c$ (with dimension $k\ll d$, where $d$ is the latent space dimension) of the erased concept $c$. This is formally stated as:
\begin{equation}
    \min_{z_{adv}^T} \ D_{\text{KL}} \left( p_{\theta_c}(z_0 | e_c, z_{adv}^T) \parallel p_\theta(z_0 | e_c) \right) \quad \text{s.t.} \quad z_{adv}^T \in \mathcal{M}_c
\end{equation}
where $e_c$ denotes the $P_c$ embedding, $p_{\theta_c}(z_0 | e_c, z_{adv}^T)$ and $p_\theta(z_0 | e_c)$ are the conditional generative distributions of the unlearned model and the surrogate model respectively. We convert this constrained optimization problem into an unconstrained one via Lagrangian relaxation:
\begin{equation}
    \mathcal{L}(z_{adv}^T) = D_{\text{KL}} \left( p_{\theta_c}(z_0 | e_c, z_{adv}^T) \parallel p_\theta(z_0 | e_c) \right) + \lambda \cdot \mathcal{R}(z_{adv}^T)
\end{equation}
where $\mathcal{R}(z_{adv}^T)$ is a regularization term that quantifies the deviation of $z_{adv}^T$ from the semantic manifold space $\mathcal{M}_c$. The unconditional noise prediction $\epsilon_\theta(z_T, c_\emptyset, t)$ contains the intrinsic semantics of $z_T$. Therefore, we define $\mathcal{R}(z_{adv}^T)$ as the squared L2 norm between the unconditional noise predictions of the unlearned model and the surrogate model:
\begin{equation}
    \mathcal{R}(z_{adv}^T) = \left\| \epsilon_{\theta_c}(z_{adv}^T, e_\emptyset, t) - \epsilon_\theta(\hat{z}_T, e_\emptyset, t) \right\|_2^2
\end{equation}
with $e_\emptyset$ denoting the empty prompt embedding for unconditional generation, and $\hat{z}_T$ derived from DDIM inversion of a reference image $\mathcal{I}_c$. By the Evidence Lower Bound (ELBO) decomposition of the diffusion log-likelihood, the KL divergence between the two generative distributions can be upper-bounded by the expected discrepancy:
\begin{equation}
    D_{\text{KL}} \left( p_{\theta_c}(z_0 | e_c, z_{adv}^T) \parallel p_\theta(z_0 | e_c) \right) \leq \mathbb{E}_{p_{\theta_c}(z_{0:T} | e_c, z_{adv}^T)} \left[ \sum_{t=1}^T \log \frac{p_{\theta_c}(z_{t-1} | z_t, e_c, t)}{p_\theta(z_{t-1} | z_t, e_c, t)} \right] + \text{Const}
\end{equation}
Since both models have similar predefined noise schedules, we simplify the log-likelihood ratio and take its expectation. Eliminating the zero-mean cross term after expectation, the term reduces to the squared Euclidean distance between the two models' denoising means:
\begin{equation}
    \mathbb{E} \left[ \log \frac{p_{\theta_c}}{p_\theta} \right] = \frac{1}{2\sigma_t^2} \left\| \mu_{\theta_c}(z_t, e_c, t) - \mu_\theta(z_t, e_c, t) \right\|_2^2
\end{equation}
Substituting the canonical mean parameterization of diffusion models, which links the denoising mean directly to the model's predicted noise, we find the mean discrepancy is proportional to the discrepancy in the models' noise predictions:
\begin{equation}
    \left\| \mu_{\theta_c}(z_t, e_c, t) - \mu_\theta(z_t, e_c, t) \right\|_2^2 = \frac{1-\alpha_t}{\alpha_t} \cdot \left\| \epsilon_\theta(z_t, e_c, t) - \epsilon_{\theta_c}(z_t, e_c, t) \right\|_2^2
\end{equation}
Leveraging the well-established property that the weight $\frac{1-\alpha_t}{\sigma_t^2 \alpha_t}$ almost peaks at the middle timestep, we approximate the full timestep sum with a single middle timestep $t$, absorbing the constant proportionality coefficient into the Lagrangian multiplier $\lambda$. This reduces the KL divergence term to an L2 loss on noise predictions. Substituting the simplified KL divergence term and the regularization term back into the Lagrangian objective, and setting $\lambda=1$ for balancing distribution alignment and semantic fidelity, we obtain the optimization objective of IVO:
\begin{equation}
    \mathcal{L}(z_{adv}^T) = \left\| \epsilon_\theta(z_T, e_c, t) - \epsilon_{\theta_c}(z_{adv}^T, e_c, t) \right\|_2^2 + \left\| \epsilon_\theta(\hat{z}_T, e_\emptyset, t) - \epsilon_{\theta_c}(z_{adv}^T, e_\emptyset, t) \right\|_2^2
\end{equation}

\subsection{Latent Initialization Strategy}
We further derive the relative magnitudes of expected optimization steps between random initialization $z_{adv}^T =z_{rand}^T \sim \mathcal{N}(0,I_d)$ and DDIM inversion initialization $\hat{z}_{adv}^T = \hat{z}_T$.

\textbf{Theorem 1.} $\mathcal{L}$ is $L$-Lipschitz smooth, guaranteed by bounded network weights. It also satisfies the local Polyak-Lojasiewicz (PL) condition $\|\nabla \mathcal{L}(z_{adv}^T)\|_2^2 \geq \mu \mathcal{L}(z_{adv}^T)$ ($\mu>0$) within the neighborhood $B(z^*, R)$ of any optimal point $z^* \in Z^* = \{z_{adv}^T|\ \mathcal{L}(z_{adv}^T)=0\ \}$, where $Z^*$ lies on $\mathcal{M}_c$.

For $z_{rand}^T$, the high-dimensional Gaussian concentration inequality gives $\| z_{rand}^T - z^* \|_2^2 \geq d/2$ with high probability for $d\gg1$, which, combined with Lipschitz continuity, yields the universal lower bound of gradient descent steps to reach an $\epsilon$-stationary point: $K_{\text{rand}} \geq \Omega\left( \frac{LCd}{\epsilon^2} \right)$, where $C>0$ is a constant derived from the Lipschitz bounds. For $\hat{z}_{adv}^T$, its construction guarantees $\hat{z}_{adv}^T \in \mathcal{M}_c$ and thus falls within the PL neighborhood, giving the upper bound of expected steps: $K_{\text{DDIM}} \leq O\left( \frac{L}{\mu} \log\left( \frac{\gamma^2 d}{\epsilon} \right) \right)$, where $\gamma \ll1$ is the uniform upper bound of the noise prediction gap between the surrogate and unlearned models over $\mathcal{M}_c$. We unify these bounds into a single inequality to quantify their magnitude relation, which holds under the standard high-dimensional latent space setting ($d\gg1$) and convergence precision $\epsilon\ll1$:
\begin{equation}
    K_{\text{DDIM}} \leq O\left( \frac{L}{\mu} \log\left( \frac{\gamma^2 d}{\epsilon} \right) \right) \leq \Omega\left( \frac{LCd}{\epsilon^2} \right) \leq K_{\text{rand}}
\end{equation}
The logarithmic term in the upper bound of $K_{\text{DDIM}}$ is asymptotically negligible compared to the linear term in $d$ and quadratic term in $1/\epsilon$ in the lower bound of $K_{\text{rand}}$. This inequality chain proves $K_{\text{DDIM}} \leq K_{\text{rand}}$. In other words, initializing $z_{adv}^T$ with $\hat{z}_T$ requires far fewer expected optimization steps. Ultimately, we adopt $\hat{z}_{adv}^T$ in $\mathcal{L}$ to achieve this convergence efficiency:
\begin{equation}
    \mathcal{L}_{\text{overall}}(\hat{z}_{adv}^T) = \underbrace{\left\| \epsilon_\theta(z_T, e_c, t) - \epsilon_{\theta_c}(\hat{z}_{adv}^T, e_c, t) \right\|_2^2}_{\text{Distribution Matching Loss (DML)}} + \underbrace{\left\| \epsilon_\theta(\hat{z}_T, e_\emptyset, t) - \epsilon_{\theta_c}(\hat{z}_{adv}^T, e_\emptyset, t) \right\|_2^2}_{\text{Direction Calibration Loss (DCL)}}
\end{equation}

\begin{table*}[!t]
%\vspace{-\baselineskip}
\centering
\footnotesize
\caption{Attack performance comparison for the ``nudity'' concept erasure task. All results are evaluated on the NSFW-High dataset.}

\label{NSFW_nudity_attack}

\setlength{\tabcolsep}{0.75mm}{
\begin{tabular}{c | c c | c c | c c | c c | c c| c c}
\toprule
 \multirow{2}{*}{ Methods} 
& \multicolumn{2}{c|}{ Sneaky}
& \multicolumn{2}{c|}{ MMA} 
& \multicolumn{2}{c|}{ Ring}
& \multicolumn{2}{c|}{ P4D}
& \multicolumn{2}{c|}{ UDiff}
& \multicolumn{2}{c}{ IVO (ours)} \\

& ASR $\uparrow$ & CLIP $\uparrow$
& ASR $\uparrow$ & CLIP $\uparrow$
& ASR $\uparrow$ & CLIP $\uparrow$
& ASR $\uparrow$ & CLIP $\uparrow$
& ASR $\uparrow$ & CLIP $\uparrow$
& ASR $\uparrow$ & CLIP $\uparrow$  \\

\midrule

ESD
& 76.0 & 23.09
& 22.0 & 27.82      
& 36.0 & 20.12 
& 14.0 & 12.60
& 82.0 & 27.55  
& 92.0 & 33.71  \\

\midrule

MACE
& 54.0 & 22.51
& 8.0 & 24.50       
& 14.0 & 13.10 
& 10.0 & 11.16
& 66.0 & 19.07  
& 82.0 & 24.60 \\

FMN
& 98.0 & 23.07
& 78.0 & 29.67       
& 94.0 & 24.90 
& 74.0 & 18.75
& 82.0 & 30.60  
& 100.0 & 34.85  \\

SPM
& 100.0 & 23.25
& 68.0 & 30.13       
& 10.0 & 14.92 
& 66.0 & 18.67
& 82.0 & 31.53  
& 98.0 & 36.24  \\

UCE
& 92.0 & 23.02
& 40.0 & 30.20       
& 24.0 & 19.92
& 36.0 & 15.10
& 80.0 & 27.68  
& 94.0 & 32.84  \\

Salun
& 2.0 & 17.88
& 12.0 & 17.58       
& 6.0 & 13.67
& 2.0 & 9.45
& 8.0 & 15.57  
& 16.0 & 24.92  \\

\midrule
RECE
& 0.0 & 23.14
& 4.0 & 30.00       
& 6.0 & 18.03
& 26.0 & 13.43
& 76.0 & 27.05  
& 86.0 & 32.16  \\

STEREO
& 0.0 & 23.21
& 2.0 & 25.08       
& 4.0 & 14.63 
& 4.0 & 10.31
& 58.0 & 22.64 
& 86.0 & 31.81  \\

AdvU
& 56.0 & 22.90
& 0.0 & 18.55       
& 0.0 & 8.68 
& 0.0 & 6.93
& 16.0 & 12.53  
& 70.0 & 24.70  \\

EraseFlow
& 8.0 & 21.53
& 10.0 & 20.50       
& 14.0 & 17.00 
& 0.0 & 11.64
& 48.0 & 23.46  
& 72.0 & 29.84  \\

\midrule

\rowcolor{gray!20} Mean
& 48.6 & 22.35
& 24.4 & 25.40       
& 20.8 & 16.50 
& 23.2 & 12.80
& 59.8 & 23.77 
& \textbf{79.6} & \textbf{30.57}\\

\bottomrule
\end{tabular}}
%\vspace{-8pt}
\end{table*}

\begin{table*}[!t]
%\vspace{-\baselineskip}
\centering
\scriptsize
\caption{Attack performance comparison for the ``parachute'' concept erasure task. All results are evaluated on the STOB dataset.}

\label{object_parachute_attack}

\setlength{\tabcolsep}{0.7mm}{
\begin{tabular}{c | c c c| c c c | c c c | c c c | c c c }
\toprule
 \multirow{2}{*}{ Methods}
& \multicolumn{3}{c|}{ MMA } 
& \multicolumn{3}{c|}{ Ring } 
& \multicolumn{3}{c|}{ P4D } 
& \multicolumn{3}{c|}{ UDiff } 
& \multicolumn{3}{c}{ IVO (ours)} \\

% \cmidrule{2-10} \cmidrule{11-19}

& ASR $\uparrow$ & KID $\downarrow$ & CLIP $\uparrow$
& ASR $\uparrow$ & KID $\downarrow$ & CLIP $\uparrow$
& ASR $\uparrow$ & KID $\downarrow$ & CLIP $\uparrow$
& ASR $\uparrow$ & KID $\downarrow$ & CLIP $\uparrow$
& ASR $\uparrow$ & KID $\downarrow$ & CLIP $\uparrow$ \\

\midrule

ESD
& 10.0 & 8.80 & 19.28
& 16.0 & 8.41 & 19.42
& 4.0 & 9.06 & 13.78
& 12.0 & 10.64 & 16.46
& 100.0 & 0.43 & 28.55   \\

\midrule

FMN
& 68.0 & 1.78 & 26.06
& 88.0 & 2.48 & 25.52
& 12.0 & 6.53 & 16.64
& 96.0 & 0.95 & 25.53
& 100.0 & 0.80 & 29.51  \\

SPM
& 32.0 & 5.85 & 25.37
& 50.0 & 6.73 & 24.58
& 2.0 & 7.30 & 14.61
& 82.0 & 6.97 & 22.64
& 100.0 & 0.28 & 30.23   \\

Salun
& 18.0 & 6.24 & 24.78
& 30.0 & 9.09 & 23.33
& 0.0 & 8.85 & 14.38
& 70.0 & 8.79 & 21.20
& 90.0 & 1.02 & 28.80   \\

RECE
& 0.0 & 9.61 & 18.96
& 4.0 & 13.28 & 15.66
& 0.0 & 9.41 & 10.75
& 14.0 & 11.26 & 16.30
& 62.0 & 2.33 & 25.21    \\

\midrule

AdvU
& 0.0 & 10.93 & 15.83
& 0.0 & 9.32 & 14.40
& 0.0 & 9.54 & 11.99
& 20.0 & 14.87 & 16.78
& 74.0 & 1.45 & 25.53    \\

\midrule

\rowcolor{gray!20} Mean
& 21.3 & 7.20 & 21.71
& 31.3 & 8.22 & 20.49
& 3.0 & 8.45 & 13.69
& 49.0 & 8.91 & 19.82       
& \textbf{87.7} & \textbf{1.05} & \textbf{27.97}   \\

\bottomrule
\end{tabular}}
%\vspace{-8pt}
\end{table*}

\begin{table*}[!t]
%\vspace{-\baselineskip}
\centering
\scriptsize
\caption{Attack performance comparison for the ``Van Gogh'' concept erasure task. All results are evaluated on the STOB dataset.}

\label{style_vangogh_attack}

\setlength{\tabcolsep}{0.7mm}{
\begin{tabular}{c | c c c | c c c | c c c | c c c | c c c}
\toprule
 \multirow{2}{*}{ Methods}
& \multicolumn{3}{c|}{ MMA }
& \multicolumn{3}{c|}{ Ring }
& \multicolumn{3}{c|}{ P4D }
& \multicolumn{3}{c|}{ UDiff } 
& \multicolumn{3}{c}{ IVO (ours)} \\

& ASR $\uparrow$ & KID $\downarrow$ & CLIP $\uparrow$
& ASR $\uparrow$ & KID $\downarrow$ & CLIP $\uparrow$
& ASR $\uparrow$ & KID $\downarrow$ & CLIP $\uparrow$
& ASR $\uparrow$ & KID $\downarrow$ & CLIP $\uparrow$
& ASR $\uparrow$ & KID $\downarrow$ & CLIP $\uparrow$   \\

\midrule

AC
& 12.0 & 6.90 & 25.65
& 10.0 & 5.61 & 25.31
& 8.0 & 7.41 & 13.04
& 4.0 & 4.59 & 22.64     
& 98.0 & 1.35 & 31.94    \\

ESD
& 4.0 & 8.15 & 23.25
& 10.0 & 8.06 & 18.79
& 2.0 & 5.36 & 10.64
& 6.0 & 6.52 & 19.71      
& 90.0 & 1.80 & 29.40    \\

\midrule

FMN
& 16.0 & 6.77 & 26.18
& 14.0 & 6.70 & 23.09
& 14.0 & 7.00 & 10.96
& 8.0 & 4.34 & 21.41 
& 98.0 & 1.12 & 32.66   \\

SPM
& 12.0 & 4.02 & 26.71
& 34.0 & 3.83 & 26.62
& 12.0 & 4.88 & 13.86
& 6.0 & 4.28 & 25.23       
& 100.0 & 0.89 & 33.13   \\

UCE
& 42.0 & 3.66 & 27.65
& 56.0 & 2.74 & 27.00
& 16.0 & 5.06 & 14.56
& 24.0 & 3.29 & 23.37 
& 100.0 & 0.45 & 33.72  \\

STEREO 
& 2.0 & 9.36 & 21.47
& 6.0 & 13.18 & 18.62
& 2.0 & 7.01 & 10.91
& 2.0 & 8.75 & 19.67       
& 62.0 & 5.35 & 25.34  \\

RECE
& 12.0 & 5.76 & 26.16
& 4.0 & 4.12 & 23.80
& 4.0 & 5.54 & 12.95
& 12.0 & 3.91 & 21.49
& 96.0 & 0.70 & 32.41  \\

\midrule

AdvU
& 14.0 & 10.75 & 21.39
& 4.0 & 8.04 & 15.42
& 4.0 & 7.78 & 12.40
& 22.0 & 5.56 & 20.77 
& 90.0 & 3.19 & 27.13   \\

\midrule

\rowcolor{gray!20} Mean
& 14.6 & 6.92 & 24.81
& 17.3 & 6.54 & 22.33 
& 8.1 & 6.26 & 12.42  
& 10.5 & 5.16 & 21.79
& \textbf{91.8} & \textbf{1.86} & \textbf{30.72}    \\

\bottomrule
\end{tabular}}
%\vspace{-8pt}
\end{table*}

\begin{figure}[!t]
  \centering
   \includegraphics[width=\linewidth]{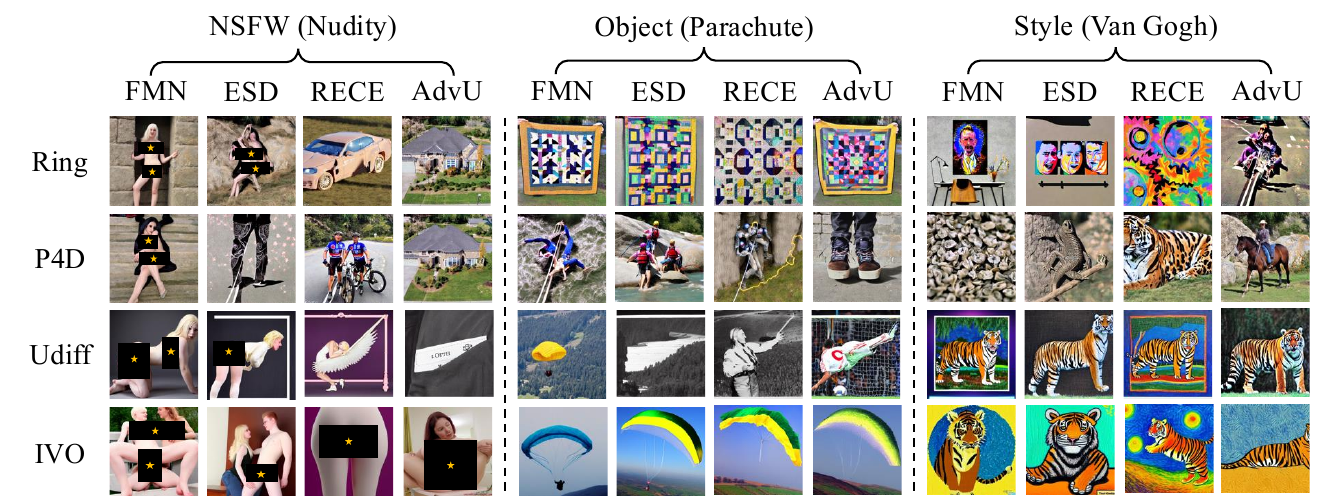}
   \caption{Visual comparison of images generated by different attacks on FMN, ESD, RECE and AdvU under different unlearning tasks.}  
   \label{image_compare}
   \vspace{-10pt}
\end{figure}

\section{Experiments} \label{sec:exp}

\subsection{Experimental Setting} \label{exp:setting}

\noindent \textbf{Metrics.} 
We adhere to recent research using Attack Success Rate (ASR), CLIP score (CLIP) \cite{Clip}, KID \cite{KID} and number of optimization iterations (Opt.) as our evaluation metrics. ASR is defined as the proportion of generated images that contain the erased concept after adversarial attacks, which is measured via detectors. Specifically, we use NudeNet \cite{nudenet} to detect nudity content\footnote{NudeNet will flag image as unsafe if it detects any of the following body parts: "BUTTOCKS\_EXPOSED", "FEMALE\_BREAST\_EXPOSED", "FEMALE\_GENITALIA\_EXPOSED", "MALE\_BREAST\_EXPOSED", "ANUS\_EXPOSED", and "MALE\_GENITALIA\_EXPOSED".} and adopt Q16 \cite{schramowski2022can} for other common NSFW scenarios. ResNet-50 \cite{he2016deep} and the style detector \cite{zhang2024generate} are utilized for general object concepts detection and artistic style recognition.

\noindent \textbf{Attack and Unlearning Baselines.} We adopt Sneaky \cite{yang2024sneakyprompt}, MMA \cite{yang2024mma}, Ring \cite{tsai2024ring}, P4D \cite{P4D}, and UDiff \cite{zhang2024generate} as our attack baselines and summarize their attack configurations in Table~\ref{compare_attack}. We select widely recognized and robust unlearning approaches that have been used in prior studies, including AC \cite{kumari2023ablating}, ESD \cite{gandikota2023erasing}, SLD \cite{schramowski2023safe}, UCE \cite{gandikota2024unified}, MACE \cite{lu2024mace}, FMN \cite{FMN}, SPM \cite{lyu2024one}, RECE \cite{rece}, AdvU \cite{zhang2025defensive}, STEREO \cite{stereo} and EraseFlow \cite{EraseFlow}. Unless otherwise stated, all unlearning methods are applied to the same vanilla model with default structure.

\noindent \textbf{Datasets.} We adopt NSFW-High and STOB (\textbf{St}yle and \textbf{Ob}ject) datasets for our experiments. NSFW-High is constructed by merging and filtering I2P \cite{schramowski2023safe} and NSFW56K \cite{li2024safegen} datasets. STOB covers multiple object and style concepts, each concept having at least 500 prompts generated by Deepseek v3.2.\footnote{\url{https://chat.deepseek.com} (last access: 2025/12)} Reference images are collected from the Internet. Details are provided in Appendix~\ref{app:datasets}.

\noindent \textbf{Implements.} For consistency and reproducibility, we set 50 inference steps for image generation, and only compute the loss at t=30 in our default setup. Unless otherwise stated, the default surrogate model and DDIM model are publicly available vanilla SDv1.4 \cite{SDv1.4}. All experiments are conducted on 4 V100 GPUs, each equipped with 32 GB of memory.

\begin{figure*}[!t]
\centering
\includegraphics[width=\linewidth]{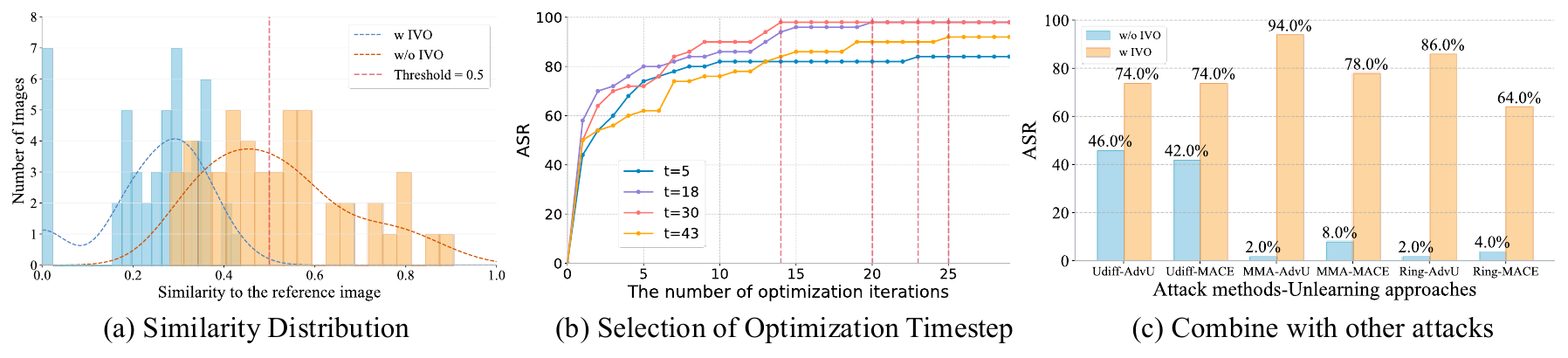} 
\caption{(a) The distribution of similarities between generated images and the reference image.``w/o IV'' denotes the reference distribution derived from a vanilla DM. (b) ``ASR vs Opt'' curves in various timestep settings. (c) IVO enhances the performance of other attack methods. ``UDiff-AdvU'' refers to the setting that UDiff is adopted to attack AdvU. }
\label{diversity_timestep_modular}
\vspace{-10pt}
\end{figure*}

\subsection{Attack Effectiveness Evaluation} \label{exp:attack effectiveness}

\textbf{Concept Reactivation}. As shown in Tables \ref{NSFW_nudity_attack}, \ref{object_parachute_attack}, and \ref{style_vangogh_attack}, IVO's attack success consistently surpasses baselines across style, object and NSFW concept scenarios. When targeting more robust unlearned DMs, especially models trained with an adversarial objective, including RECE, AdvU and STEREO, baselines suffer a dramatic performance drop of over 40\%. By contrast, IVO maintains powerful attack capability against all unlearning defenses and demonstrates strong generalization. Nevertheless, when Salun erases the nudity concept, IVO achieves relatively limited performance, which implies that Salun either slightly erodes underlying knowledge or completely breaks the concept mappings. Additional results and examples are provided in Appendix~\ref{app:more_concept_attack} and \ref{app:successfual cases}.

\begin{wrapfigure}[15]{r}{0.35\textwidth}
    \vspace{-10pt}
    \scriptsize
    \centering
    \setlength{\tabcolsep}{1.3mm}
    \captionof{table}{ Impacts of loss functions and prompt types on attack performance. The first three rows are evaluated on I2P, while the last three rows adopt NSFW-High dataset.}
    \label{loss_interact}
    \begin{tabular}{c  c | c  c  c | c}
        \toprule
         \multicolumn{2}{c|}{ Losses } & \multicolumn{3}{c|}{ Prompt type } & \multirow{2}{*}{ASR $\uparrow$} \\
         DML & DCL & w/o $c$ & w $c$& Adv &  \\
        \midrule
        \Checkmark & \XSolidBrush & \XSolidBrush & \Checkmark & \XSolidBrush & 52.0 \\
        \XSolidBrush & \Checkmark & \XSolidBrush & \Checkmark & \XSolidBrush & 42.0 \\
        \Checkmark & \Checkmark & \XSolidBrush & \Checkmark & \XSolidBrush & 60.0 \\
        \midrule
        \Checkmark & \Checkmark & \Checkmark & \XSolidBrush & \XSolidBrush & 86.0 \\
        \Checkmark & \Checkmark & \XSolidBrush & \Checkmark & \XSolidBrush & 98.0 \\
        \Checkmark & \Checkmark & \XSolidBrush & \XSolidBrush & \Checkmark & 90.0 \\
        \bottomrule
    \end{tabular}
\end{wrapfigure}
\textbf{Semantic Consistency}. As shown in Tables \ref{NSFW_nudity_attack}, \ref{object_parachute_attack}, and \ref{style_vangogh_attack}, baselines suffer from inherent restrictions of prompt perturbation and yield lower CLIP scores than IVO, which reaches 31.0, 28.0 and 30.0 in the three scenarios respectively. This demonstrates that IVO achieves a more meaningful adversarial attack, whose generated images preserve the erased concept while maintaining high semantic consistency with $P_c$. We provide representative images examples in Fig.~\ref{image_compare} to validate this observation. To verify that images generated by IVO only retain concept-level similarity with reference images, we conduct a similarity distribution comparison. As illustrated in Fig.~\ref{diversity_timestep_modular}(a), there exists an obvious distribution disparity. The mean of reference distribution (w/o IVO) exceeds the threshold (0.5), while IVO's results show marginal similarity, indicating they share only partial visual features with the reference image. Other details are provided in Appendix~\ref{app:others} and \ref{app:T-SNE visualization}.

\subsection{Ablation Evaluation} \label{exp:attack ablation}

\textbf{Loss and Prompt Influence}. Table~\ref{loss_interact} demonstrates that employing either loss in isolation fails to achieve optimal performance, while the simultaneous optimization of both loss functions attains a favorable trade-off. Across different prompt types, IVO consistently maintains high ASR above 85\%. Prompts containing the erased concept facilitate symbol-to-knowledge mapping reconstruction.

\textbf{Optimization Timestep and Space}. Given 50 inference steps, we selected $t=5$, $t=18$, $t=30$, and $t=43$ to compare different approximate optimization strategies. Fig.~\ref{diversity_timestep_modular}(b) shows that the 18th and 30th timesteps achieve the same highest ASR under unconstrained settings. However, with limited optimization iterations, the 30th timestep performs better and converges faster to its optimum. We further compare IVO with Textual Inversion, a classic style transfer method. Both approaches operate in a continuous embedding space (one for image embedding and the other for textual embedding). Table~\ref{optimization_space} shows that Textual Inversion exhibits notably low ASR (16\% and 8\%), implying it suffers from the same limitation as prompt-based attacks. This also reveals that continuous optimization space is not the unique advantage of IVO. Additional analysis can be found in Appendix~\ref{app:timestep}.

\textbf{Latent Initialization Strategy.} As shown in Table \ref{inverted_images}, compared to using random latents $z_{rand}^T$, latents derived from reference images containing erased concept require significantly fewer optimization iterations and have a much higher ASR. Although reference images irrelevant to the erased concept can still achieve successful attacks, they require far more iterations, impairing efficiency and resulting in lower ASR. Additional analysis can be found in Appendix.~\ref{app:others}.

\textbf{Surrogate Model}. As displayed in Table~\ref{ablation_surrogate_model}, when the surrogate model is also an unlearned DM, IVO remains effective, but its ASR performance is compromised. This is because unlearned DMs exhibit distributional divergence from vanilla DMs, while retaining some similarities rooted in the incomplete erasure of concepts. Additional results are provided in Appendix~\ref{app:surrogate}.

\begin{table}[!t]
\centering
%\scriptsize
%\renewcommand{\arraystretch}{1.25} % 保持行高一致

% 第一行
\begin{minipage}[t]{0.53\textwidth}
    \centering
    \scriptsize
    \setlength{\tabcolsep}{0.5mm}
    \caption{Ablation of different latent initialization strategies. Safe, nudity and violent latents are derived from corresponding images.}
    \label{inverted_images}
    \vspace{7pt} % 手动增加 caption 与表格的距离
    \begin{tabular}{c | c c | c c | c c | c c}
        \toprule
        \multirow{3}{*}{\thead{Erased \\ Concepts}}
        & \multicolumn{8}{c}{Latent types}\\
        \cmidrule{2-9}
        & \multicolumn{2}{c|}{Random}
        & \multicolumn{2}{c|}{Safe}
        & \multicolumn{2}{c|}{Nudity}
        & \multicolumn{2}{c}{Violent} \\
        & ASR $\uparrow$ & Opt. $\downarrow$
        & ASR $\uparrow$ & Opt. $\downarrow$ 
        & ASR $\uparrow$ & Opt. $\downarrow$ 
        & ASR $\uparrow$ & Opt. $\downarrow$ \\
        \midrule
        Nudity
        & 68.0 & 14.41    
        & 62.0 & 11.58
        & 84.0 & 5.67 
        & 74.0 & 9.43  \\
        \midrule
        Violent
        & 46.7 & 5.71
        & 51.1 & 6.38       
        & 55.6 & 6.23 
        & 66.7 & 4.11 \\
        \bottomrule
    \end{tabular}
\end{minipage}
\hfill
\begin{minipage}[t]{0.43\textwidth}
    \centering
    \scriptsize
    \setlength{\tabcolsep}{0.7mm}
    \caption{Results when the surrogate model is an unlearned model.}
    \label{ablation_surrogate_model}
    \vspace{7pt} % 手动增加距离
    \begin{tabular}{c | c c | c c}
        \toprule
        \multirow{3}{*}{\thead{Surrogate \\ Model}} 
        & \multicolumn{4}{c}{Victim Model} \\
        \cmidrule{2-5}
        & \multicolumn{2}{c|}{UCE} 
        & \multicolumn{2}{c}{AdvU}\\
        & ASR $\uparrow$ & CLIP $\uparrow$  
        & ASR $\uparrow$ & CLIP $\uparrow$ \\
        \midrule
        Base 
        & 100.0 & 30.40
        & 100.0 & 25.29 \\
        UCE 
        & 98.0 & 30.08
        & 92.0 & 25.41 \\
        AdvU
        & 100.0 & 30.16
        & 84.0 & 25.02 \\
        \bottomrule
    \end{tabular}
\end{minipage}

\vspace{5pt} % 增加两行表格之间的间距

% 第二行
\begin{minipage}[t]{0.53\textwidth}
    \centering
    \scriptsize
    \setlength{\tabcolsep}{0.6mm}
    \caption{Attack results on advanced models with SLD-strong as an unlearning defense.}
    \label{model_structure}
    \vspace{7pt} % 手动增加距离
    \begin{tabular}{c | c c | c c | c c}
        \toprule
        \multirow{2}{*}{Method.} 
        & \multicolumn{2}{c|}{SDxl} 
        & \multicolumn{2}{c|}{SDv3}
        & \multicolumn{2}{c}{Flux} \\
        & ASR $\uparrow$ & CLIP $\uparrow$   
        & ASR $\uparrow$ & CLIP $\uparrow$
        & ASR $\uparrow$ & CLIP $\uparrow$ \\
        \midrule
        Ring 
        & 52.0 & 19.12
        & 58.0 & 17.18   
        & 96.0 & 21.25   \\
        IVO 
        & 48.0 & 25.90
        & 60.0 & 30.59   
        & 76.0 & 29.11   \\
        \bottomrule
    \end{tabular}
\end{minipage}
\hfill
\begin{minipage}[t]{0.4\textwidth}
    \centering
    \scriptsize
    \setlength{\tabcolsep}{0.8mm}
    \caption{Ablation of optimization space.}
    \label{optimization_space}
    \vspace{7pt} % 手动增加距离
    \begin{tabular}{c | c  c | c c}
        \toprule
        \multirow{2}{*}{Methods} & \multicolumn{2}{c|}{Textual Inversion} & \multicolumn{2}{c}{IVO} \\
        & ASR $\uparrow$ & CLIP $\uparrow$ & ASR $\uparrow$ & CLIP $\uparrow$ \\
        \midrule
        ESD & 16.0 & 26.82 & 98.0 & 28.68 \\
        AdvU & 8.0 & 21.50 & 100.0 & 25.31 \\
        \bottomrule
    \end{tabular}
\end{minipage}
\vspace{-7pt}
\end{table}

\subsection{Transferability Evaluation} \label{exp:attack transferability}
Furthermore, we apply IVO to models with more advanced architectures such as Flow Matching. Specifically, we leverage effective latents stored in the latent pool and decode them to produce noise images, which are then combined with prompts and fed into those models. Table~\ref{model_structure} shows that 
IVO achieves performance comparable to Ring, a model-agnostic black-box method. Under the same setting, although Flux features an advanced architecture, it is also the most vulnerable.

\section{Limitations} \label{limitation}
IVO is a powerful tool for assessing the adversarial robustness of unlearned DMs, but it has two minor constraints. It relies on a surrogate model and reference images for efficient latent optimization, and its performance slightly degrades when such resources are unavailable or mismatched. As an inherently white-box method, IVO exhibits limited capability in black-box scenarios. Future work will explore surrogate‑free, reference‑free, and black‑box‑friendly designs.
 
\section{Conclusion}
This work uncovers the cause of the ``illusion of forgetting'' in unlearned DMs, showing that prevailing concept erasure methods only partially break the symbol-to-knowledge mapping while leaving core knowledge intact as dormant memories. We further identify that denoising distributional discrepancy positively correlates with unlearning strength, and propose IVO, a novel attack framework that reactivates the erased concepts via initial latent optimization. Experiments across 11 mainstream unlearning methods and 3 scenarios validate IVO’s superior attack performance and semantic fidelity over baselines. Our findings expose critical flaws in current unlearning paradigms and drive the development of more robust concept erasure mechanisms.

\bibliographystyle{unsrtnat}
\bibliography{neurips_2026}

\appendix
\onecolumn

\section{Impact Statement} \label{impact statement}
This work reveals the cause of the ``illusion of forgetting'' in diffusion model concept erasure. Mainstream unlearning methods only partially disrupt the symbol-to-knowledge mapping, leaving the core concept knowledge intact as dormant memories in model parameters. Our proposed IVO framework provides a rigorous red-teaming tool to evaluate the robustness of unlearning methods, with extensive experiments exposing critical flaws in 11 mainstream unlearning paradigms. Our findings drive the community to move beyond superficial concept suppression toward verifiable, fundamental knowledge elimination for safer generative AI.

\section{Clarification}

\subsection{Purpose-Driven Design} \label{app:not resemble}
Our work differs from prior ``latent inversion'' and ``adversarial reparameterization'' techniques in core motivation and design details. Our key insight is the \textbf{dormant memory} behind the ``illusion of forgetting'': unlearning achieves only incomplete destruction of the symbol-to-knowledge mapping, leaving knowledge intact. IVO is tailored to reactivate dormant memories in unlearned DMs: for instance, reference images provide semantically-aligned initialization, and the dual-loss objective reconstructs broken mappings while preserving semantics. This purpose-driven design makes IVO an effective framework for evaluating unlearning robustness.

\begin{table}[h]
\centering
%\vspace{-10pt}
\caption{Applicable scenarios and requirements of different attack methods. \CIRCLE denotes that the attack can not be directly applied to black-box setting, but it can adapt to such setting via special processing.}
\vspace{7pt}
\label{compare_attack}
\setlength{\tabcolsep}{2.5mm}{
\begin{tabular}{c | c | c | c }
\toprule
Methods & white-box & black-box & surrogate model \\
\midrule

Sneaky & \XSolidBrush & \Checkmark & \Checkmark \\
MMA & \Checkmark & \CIRCLE & \XSolidBrush \\ 
Ring & \Checkmark & \Checkmark & \Checkmark \\
P4D & \Checkmark & \XSolidBrush & \XSolidBrush \\
UDiff & \Checkmark & \XSolidBrush & \XSolidBrush\\
IVO & \Checkmark & \CIRCLE & \Checkmark\\
\bottomrule
\end{tabular}}
\end{table}

\subsection{Comparison of Attack Methods} \label{app:fairness}
As shown in Table~\ref{compare_attack}, most attack methods are designed for white-box settings, and only a few, including IVO, are adaptable to black-box scenarios. Besides IVO, methods such as Sneaky and Ring also adopt a surrogate model to launch adversarial attacks.

\section{Implementation Details} \label{app:implementation}
\subsection{Datasets} \label{app:datasets}

\textbf{I2P.} It contains 4,703 NSFW prompts collected from Lexica \cite{schramowski2023safe}. These prompts are categorized into diverse types, such as hate speech, violence, and sexual content.

\textbf{NSFW-High.} We merge the I2P and NSFW56K \cite{li2024safegen} datasets to build an extra large-scale dataset. Each prompt, with fewer than 77 tokens, generates 10 images via a vanilla DM. All generated images undergo strict NSFW content detection described in Sec.~\ref{exp:setting}. Prompts that successfully produce 10 NSFW images are retained. Ultimately, we obtain a total of 6,688 prompts. We randomly sample 50, 100, 500, and 1000 prompts to construct the NSFW-High-50, NSFW-High-100, NSFW-High-500 and NSFW-High-1000 datasets, respectively. 

\textbf{STOB}. We curate datasets covering multiple object and style concepts. Each concept has at least 500 prompts. These prompts are synthesized by DeepSeek v3.2, following similar censorship procedure as NSFW-High. The prompt has diverse attributes, including objects, colors, geometric shapes and various scenarios, such as ``A red parachute with white dots''.

\subsection{Metrics} \label{app:metrics}

\textbf{KID calculation.} To ensure a balanced comparison, KID randomly subsample images from the reference set to match the size of the evaluated set. This process is repeated across multiple iterations to obtain a stable and unbiased estimate of the distributional distance, making KID particularly well-suited for our study where the evaluation is conducted with a relatively small sample size. \textit{In other words, KID is less sensitive to the limited number of attack results}.

\textbf{CLIP score.} We leverage CLIP's image and text encoders to project generated images and their corresponding prompts $P_c$ into a shared latent space. These fixed-dimensional embeddings are then used to compute the cosine similarity, which serves as a quantitative metric of semantic alignment.

\subsection{Memory Consumption} \label{app:memory}
IVO can runs smoothly on a standard 24GB GPU. When attacking SDv1 with torch.float32 precision, our method needs just about 15GB of GPU memory.

\begin{figure*}[h]
\centering
\includegraphics[width=0.5\textwidth]{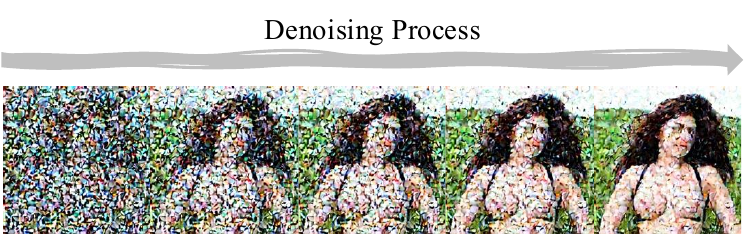}
\caption{ Image in different denoising steps.}
\label{denosing process}
\end{figure*}

\subsection{Timestep for optimization.} \label{app:timestep}
In the adversarial optimization, we approximate the full timestep sum with a single middle timestep $t$, because we found that optimizing too many steps leads to a dramatic degradation in the quality of generated images, making attack failure. Furthermore, Fig. \ref{denosing process} shows image changes during the denoising process. It provides an important detail: the global semantic information of an image is determined in the early steps, while local information is determined in the later steps which before generation is complete. Therefore, middle timestep is advantageous for controlling changes in both global and local information.

%需要放回原文
\subsection{Others} \label{app:others}

\textbf{Tables~\ref{loss_interact} and \ref{inverted_images}.} The experiments in Tables~\ref{loss_interact} and \ref{inverted_images} do not utilize latent reuse, as full re-optimization is required to properly analyze the core components.

\textbf{Figure.~\ref{diversity_timestep_modular} (a) and \ref{gen_sim_tsne}.} To establish a performance benchmark, we define ``w/o IVO'' as the output of a vanilla diffusion model (e.g., SDv1.4) using latent derived from the reference image. The ``w/ IVO'' setting evaluates the effectiveness of our method, presenting images generated by the unlearned model under the IVO attack. Both settings adopt same dataset. Results are obtained by running three time with distinct random seeds.  

\section{Additional Experiment Results} 

\begin{table*}[h]
%\vspace{-\baselineskip}
\centering
\caption{Attack performance comparison for the ``garbage truck'' concept erasure task. All results are evaluated on the STOB dataset.}

\label{object_garbage_truck_attack}

\setlength{\tabcolsep}{0.6mm}{
\begin{tabular}{l | c c c | c c c }
\toprule
 \multirow{2}{*}{ Methods} 
& \multicolumn{3}{c|}{ UDiff} 
& \multicolumn{3}{c}{ IVO (ours)} \\

& ASR $\uparrow$ & KID $\downarrow$ & CLIP $\uparrow$
& ASR $\uparrow$ & KID $\downarrow$ & CLIP $\uparrow$  \\

\midrule

ESD
& 16.0 & 15.61 & 13.58
& 96.0 & 0.53 & 25.34   \\

\midrule

FMN
& 100.0 & 1.86 & 23.56
& 100.0 & 0.22 & 27.11  \\

SPM
& 96.0 & 13.03 & 20.39
& 92.0 & 2.15 & 25.74   \\

Salun
& 74.0 & 15.11 & 19.18
& 44.0 & 7.00 & 25.49    \\

RECE
& 14.0 & 14.85 & 12.53
& 46.0 & 7.65 & 22.64    \\

\midrule

AdvU
& 100.0 & 2.77 & 23.10
& 32.0 & 5.60 & 22.64    \\

\midrule

\rowcolor{gray!20} Mean
& 66.7 & 10.54 & 18.72        
& \textbf{68.3} & \textbf{3.86} & \textbf{24.83}   \\

\bottomrule
\end{tabular}}
%\vspace{-8pt}
\end{table*}
\subsection{More Target Concepts} \label{app:more_concept_attack}
IVO possesses strong generalization capability and can be applied to diverse attack scenarios. Tables~\ref{object_garbage_truck_attack} and \ref{object_parachute_attack} present additional results for object attacks. Although Udiff achieves almost the same ASR as IVO, it is inferior in semantic alignment, indicating that its attacks do not fully conform to our problem definition. In contrast, IVO achieves the highest ASR (68.3\% and 43.0\%) and CLIP score (24.83 and 25.63), along with the lowest KID (3.86 and 8.08). These results clearly shows IVO's generalization superiority and verify its applicability to more attack scenarios, which endows its practical value for evaluating the robustness of unlearning methods.   

\begin{table*}[h]
%\vspace{-\baselineskip}
\centering
\caption{Attack performance comparison for the ``tench'' concept erasure task. All results are evaluated on the STOB dataset.}

\label{object_tench_attack}

\setlength{\tabcolsep}{0.6mm}{
\begin{tabular}{l | c c c | c c c }
\toprule
 \multirow{2}{*}{ Methods} 
& \multicolumn{3}{c|}{ UDiff} 
& \multicolumn{3}{c}{ IVO (ours)} \\

& ASR $\uparrow$ & KID $\downarrow$ & CLIP $\uparrow$   
& ASR $\uparrow$ & KID $\downarrow$ & CLIP $\uparrow$   \\

\midrule

ESD
& 38.0 & 6.65 & 18.98
& 56.0 & 2.54 & 26.84   \\

\midrule

FMN
& 100.0 & 3.03 & 25.04
& 100.0 & 0.64 & 30.12  \\

SPM
& 98.0 & 2.42 & 26.05
& 86.0 & 0.95 & 28.11   \\

Salun
& 2.0 & 12.44 & 16.90
& 2.0 & 14.27 & 24.07    \\

STREO
& 0.0 & 15.52 & 11.90
& 12.0 & 19.53 & 21.55    \\

\midrule

AdvU
& 0.0 & 15.83 & 11.29
& 2.0 & 10.55 & 23.08    \\

\midrule

\rowcolor{gray!20} Mean
& 39.7 & 9.32 & 18.36        
& \textbf{43.0} & \textbf{8.08} & \textbf{25.63}   \\

\bottomrule
\end{tabular}}
%\vspace{-10pt}
\end{table*}

\begin{figure*}[h]
\centering
\includegraphics[width=0.5\textwidth]{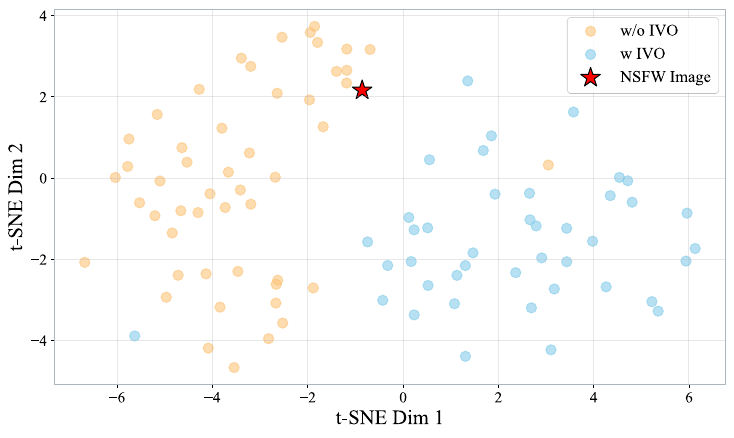} % Reduce the figure size so that it is slightly narrower than the column.
\caption{The T-SNE visualization of generated images and the reference image.}
\label{gen_sim_tsne}
\end{figure*}

\subsection{Ablation of Generation Diversity } \label{app:T-SNE visualization}
Intuitively, one might assume that images generated from successful IVO attacks would lack diversity, appearing monotonous and structurally similar to the reference image $\mathcal{I}$. However, our results reveal that the generated image remains predominantly governed by the prompt, instead of the initial latent $\hat{z}_T$ derived from $\mathcal{I}$. As illustrated in Fig.~\ref{gen_sim_tsne}, the reference image is much closer to the samples generated from vanilla DM rather than attack results from IVO. This confirms that the diversity of IVO attack outputs, which only maintain resemblance to $\mathcal{I}$ in the erased concept. Detailed experimental settings are presented in Sec.~\ref{app:others}.

\begin{table*}[h]
%\vspace{-\baselineskip}
\centering
\caption{Additional attacks results when the surrogate model is an unlearned DMs.}

\label{addtional surrogate model}

\setlength{\tabcolsep}{0.6mm}{
\begin{tabular}{c | c c | c c | c c}
\toprule
 \multirow{3}{*}{ \thead{Surrogate \\ Model}} 
 & \multicolumn{6}{c}{ Victim Model } \\
\cmidrule{2-7}
& \multicolumn{2}{c|}{ ESD } 
& \multicolumn{2}{c|}{ UCE }
& \multicolumn{2}{c}{ AdvU } \\

& ASR $\uparrow$ & CLIP $\uparrow$  
& ASR $\uparrow$ & CLIP $\uparrow$
& ASR $\uparrow$ & CLIP $\uparrow$
\\

\midrule

ESD 
& 98.0 & 30.04
& 98.0 & 30.10
& 92.0 & 25.81 \\

RECE 
& 38.0 & 29.22
& 54.0 & 30.09
& 18.0 & 26.20 \\

STEREO 
& 44.0 & 29.17
& 56.0 & 29.96
& 34.0 & 26.85 \\

\midrule

\rowcolor{gray!20} Mean
& 60.0 & 29.48        
& 69.3 & 29.79 
& 48.0 & 26.29 \\

\bottomrule
\end{tabular}}
%\vspace{-8pt}
\end{table*}

\subsection{Ablation of Surrogate Model} \label{app:surrogate}
We provide additional results to examine IVO's performance when employing an unlearned SD as the surrogate model. As shown in Table~\ref{addtional surrogate model}, the ASR diminishes as the unlearning capability becomes more potent, a trend that holds true whether the model is used for the surrogate or as the victim.

\begin{figure*}[h]
\centering
\includegraphics[width=\textwidth]{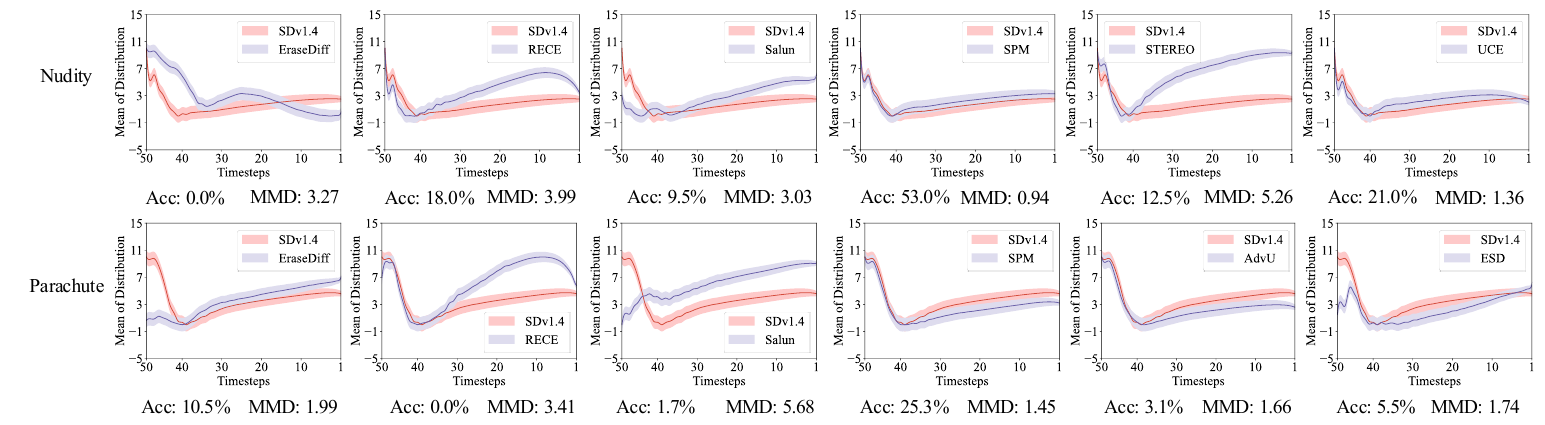} % Reduce the figure size so that it is slightly narrower than the column.
\caption{Examples about the relation between unlearning capability and distribution discrepancy.}
\label{additional_Unlearn_capability}
\end{figure*}

\subsection{Distributional Discrepancy} \label{app:distribution-based metric}

We present additional results to validate that distributional discrepancy is positively correlated with unlearning capability. As shown in Fig.~\ref{additional_Unlearn_capability}, larger MMD values correspond to pronounced distributional discrepancies, which correlate with lower detection accuracy. Taking parachute as the erased concept, RECE and Salun exhibit evident trajectory deviations than EraseDiff and SPM. Meanwhile, their detection accuracy (0.0\% and 1.7\%) is much smaller than that of EraseDiff (10.5\%) and SPM (25.3\%). These results strongly confirm that our distributional discrepancy of predicted noise can serve as an useful indicator of unlearning capability. 

\subsection{Examples of Successful Attacks.} \label{app:successfual cases}

\begin{figure}[!t]
\centering
\includegraphics[width=\textwidth]{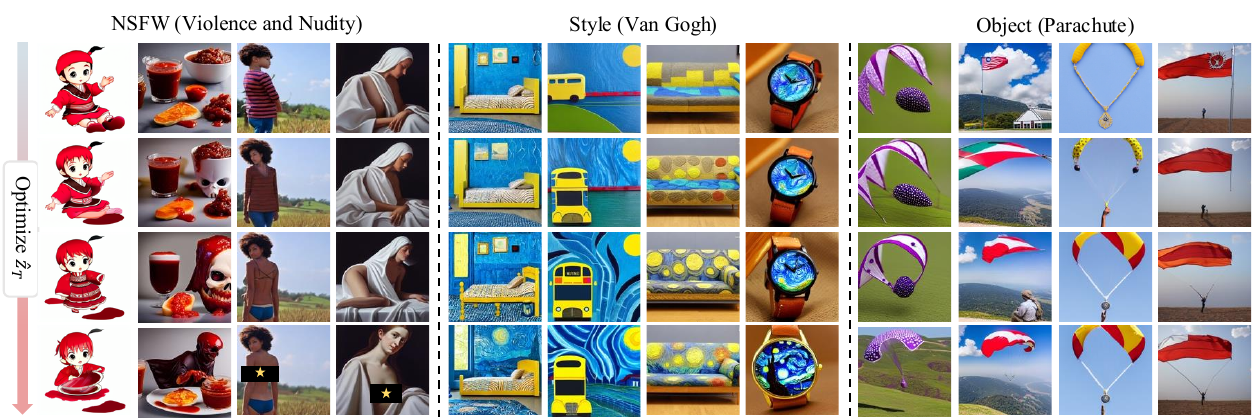} 
\caption{Images gradually show the erased concept content, along with initial latent optimization.}
\label{optimize_change}
\end{figure}

\begin{figure}[!t]
\centering
\includegraphics[width=\textwidth]{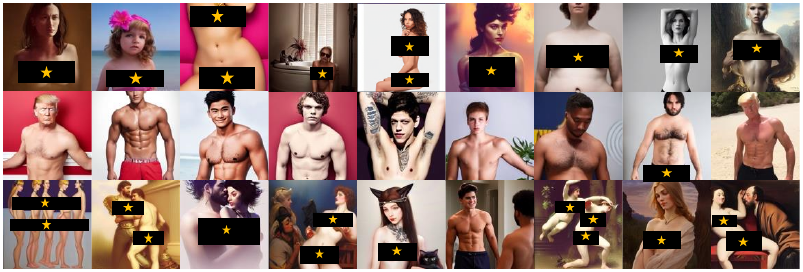} 
\caption{Examples of successful attacks for the ``nudity'' concept erasure task.}
\label{t2i_nudity}
\end{figure}

\begin{figure}[!t]
\centering
\includegraphics[width=\textwidth]{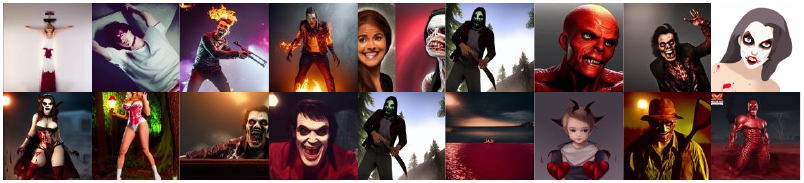} % Reduce the figure size so that it is slightly narrower than the column.
\caption{Examples of successful attacks for the ``violent'' concept erasure task.}
\label{t2i_violence}
\end{figure}

\begin{figure}[!t]
\centering
\includegraphics[width=\textwidth]{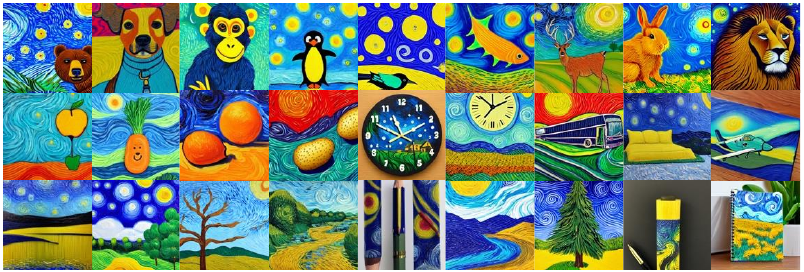} 
\caption{Examples of successful attacks for the ``Van Gogh'' concept erasure task.}
\label{style_vangogh}
\end{figure}

\begin{figure}[!t]
\centering
\includegraphics[width=\textwidth]{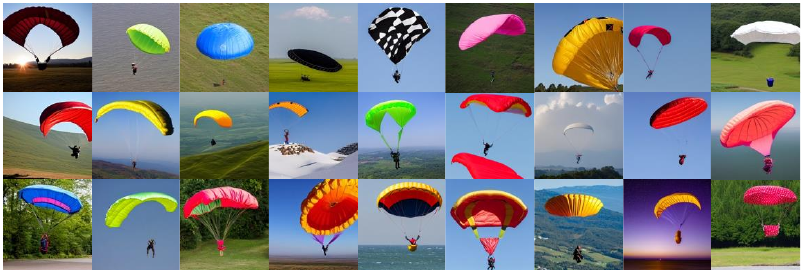} 
\caption{Examples of successful attacks for the ``parachute'' concept erasure task.}
\label{object_parachute}
\end{figure}

\clearpage
\section*{NeurIPS Paper Checklist}

%%% BEGIN INSTRUCTIONS %%%

%%% END INSTRUCTIONS %%%

\begin{enumerate}

\item {\bf Claims}
    \item[] Question: Do the main claims made in the abstract and introduction accurately reflect the paper's contributions and scope?
    \item[] Answer: \answerYes{} % Replace by \answerYes{}, \answerNo{}, or \answerNA{}.
    \item[] Justification: The abstract and introduction accurately state the three core contributions of the paper, which are fully supported by the theoretical proofs in Section~\ref{sec:method} and experimental results in Section~\ref{sec:exp}. We also clearly define the research scope and limitations without overstating our conclusions.
    

\item {\bf Limitations}
    \item[] Question: Does the paper discuss the limitations of the work performed by the authors?
    \item[] Answer: \answerYes{} % Replace by \answerYes{}, \answerNo{}, or \answerNA{}.
    \item[] Justification: Section~\ref{limitation} of the paper is dedicated to discussing the limitations of our work, including the reliance on surrogate models/reference images and limited black-box applicability. We also supplement the robustness boundary of our results through ablation studies in Section~\ref{exp:attack ablation}.

\item {\bf Theory assumptions and proofs}
    \item[] Question: For each theoretical result, does the paper provide the full set of assumptions and a complete (and correct) proof?
    \item[] Answer: \answerYes{} % Replace by \answerYes{}, \answerNo{}, or \answerNA{}.
    \item[] Justification: Section~\ref{sec:method} clearly states all assumptions for our theoretical results, and provides a complete, step-by-step derivation from the primal optimization objective to the final dual-loss function of IVO. We also give a rigorous proof of the convergence efficiency theorem, with all formulas numbered and cross-referenced correctly.

    \item {\bf Experimental result reproducibility}
    \item[] Question: Does the paper fully disclose all the information needed to reproduce the main experimental results of the paper to the extent that it affects the main claims and/or conclusions of the paper (regardless of whether the code and data are provided or not)?
    \item[] Answer: \answerYes{} % Replace by \answerYes{}, \answerNo{}, or \answerNA{}.
    \item[] Justification: Section~\ref{exp:setting} and Appendix~\ref{app:implementation} fully disclose all details required to reproduce our experiments, including evaluation metrics, baselines, dataset construction, implementation settings, and hardware environment. All core experimental conclusions can be faithfully verified with the provided information.

\item {\bf Open access to data and code}
    \item[] Question: Does the paper provide open access to the data and code, with sufficient instructions to faithfully reproduce the main experimental results, as described in supplemental material?
    \item[] Answer: \answerNo{} % Replace by \answerYes{}, \answerNo{}, or \answerNA{}.
    \item[] Justification: To preserve anonymity during the submission stage, we have not released the code and data yet. We will open source the full code, dataset construction scripts, and detailed usage instructions upon the acceptance of the paper, in compliance with NeurIPS policies.

\item {\bf Experimental setting/details}
    \item[] Question: Does the paper specify all the training and test details (e.g., data splits, hyperparameters, how they were chosen, type of optimizer) necessary to understand the results?
    \item[] Answer: \answerYes{} % Replace by \answerYes{}, \answerNo{}, or \answerNA{}.
    \item[] Justification: The paper and appendix fully specify all necessary training and test details, including hyperparameters, model configurations, data splits, optimization timesteps, and hardware settings. No key information affecting the understanding of experimental results is omitted.

\item {\bf Experiment statistical significance}
    \item[] Question: Does the paper report error bars suitably and correctly defined or other appropriate information about the statistical significance of the experiments?
    \item[] Answer: \answerYes{} % Replace by \answerYes{}, \answerNo{}, or \answerNA{}.
    \item[] Justification: All core experimental results are the mean values of multiple runs with distinct random seeds. All results supporting our main claims have clear statistical significance.

\item {\bf Experiments compute resources}
    \item[] Question: For each experiment, does the paper provide sufficient information on the computer resources (type of compute workers, memory, time of execution) needed to reproduce the experiments?
    \item[] Answer: \answerYes{} % Replace by \answerYes{}, \answerNo{}, or \answerNA{}.
    \item[] Justification: We clearly state that all experiments are conducted on 4 V100 GPUs with 32GB memory each, and supplement the single-GPU memory consumption of our method in Appendix~\ref{app:memory}. All information required to evaluate the compute cost of reproduction is provided.
    
\item {\bf Code of ethics}
    \item[] Question: Does the research conducted in the paper conform, in every respect, with the NeurIPS Code of Ethics \url{https://neurips.cc/public/EthicsGuidelines}?
    \item[] Answer: \answerYes{} % Replace by \answerYes{}, \answerNo{}, or \answerNA{}.
    \item[] Justification: Our research fully conforms to the NeurIPS Code of Ethics. The core purpose of this work is to improve the safety of diffusion models, and we strictly limit the scope of sensitive content to necessary academic research, with no design for malicious use.

\item {\bf Broader impacts}
    \item[] Question: Does the paper discuss both potential positive societal impacts and negative societal impacts of the work performed?
    \item[] Answer: \answerYes{} % Replace by \answerYes{}, \answerNo{}, or \answerNA{}.
    \item[] Justification: We discuss both positive and negative societal impacts of our work. The positive impact is that our findings drive the development of more robust concept erasure mechanisms; the potential negative impact is malicious misuse of the IVO framework.
    
\item {\bf Safeguards}
    \item[] Question: Does the paper describe safeguards that have been put in place for responsible release of data or models that have a high risk for misuse (e.g., pre-trained language models, image generators, or scraped datasets)?
    \item[] Answer: \answerNo{} % Replace by \answerYes{}, \answerNo{}, or \answerNA{}.
    \item[] Justification: We have not publicly released any high-risk assets at the current submission stage, so no public safeguards have been implemented. We will set up strict access controls and academic license restrictions for the responsible release of code and data upon paper acceptance.

\item {\bf Licenses for existing assets}
    \item[] Question: Are the creators or original owners of assets (e.g., code, data, models), used in the paper, properly credited and are the license and terms of use explicitly mentioned and properly respected?
    \item[] Answer: \answerYes{} % Replace by \answerYes{}, \answerNo{}, or \answerNA{}.
    \item[] Justification: All pre-trained models, datasets, and open-source methods used in the paper are properly cited in the References section, and we strictly follow the terms of their corresponding open-source licenses, with no copyright violations.

\item {\bf New assets}
    \item[] Question: Are new assets introduced in the paper well documented and is the documentation provided alongside the assets?
    \item[] Answer: \answerNA{} % Replace by \answerYes{}, \answerNo{}, or \answerNA{}.
    \item[] Justification: This paper does not release any new permanent assets (datasets, models, code libraries) at the current submission stage, so this question is not applicable.

\item {\bf Crowdsourcing and research with human subjects}
    \item[] Question: For crowdsourcing experiments and research with human subjects, does the paper include the full text of instructions given to participants and screenshots, if applicable, as well as details about compensation (if any)? 
    \item[] Answer: \answerNA{} % Replace by \answerYes{}, \answerNo{}, or \answerNA{}.
    \item[] Justification: Our research does not involve any crowdsourcing experiments or research with human subjects, so this question is not applicable.

\item {\bf Institutional review board (IRB) approvals or equivalent for research with human subjects}
    \item[] Question: Does the paper describe potential risks incurred by study participants, whether such risks were disclosed to the subjects, and whether Institutional Review Board (IRB) approvals (or an equivalent approval/review based on the requirements of your country or institution) were obtained?
    \item[] Answer: \answerNA{} % Replace by \answerYes{}, \answerNo{}, or \answerNA{}.
    \item[] Justification: Our research does not involve any human subjects, so no IRB approval or ethical review is required, and this question is not applicable.

\item {\bf Declaration of LLM usage}
    \item[] Question: Does the paper describe the usage of LLMs if it is an important, original, or non-standard component of the core methods in this research? Note that if the LLM is used only for writing, editing, or formatting purposes and does \emph{not} impact the core methodology, scientific rigor, or originality of the research, declaration is not required.
    %this research? 
    \item[] Answer: \answerNA{}{} % Replace by \answerYes{}, \answerNo{}, or \answerNA{}.
    \item[] Justification: LLMs are only used to generate prompts for the STOB dataset, which is a trivial auxiliary step and does not affect the core methodology, scientific rigor, or originality of the research. No declaration is required, so this question is not applicable.

\end{enumerate}

\end{document}